\documentclass[10pt,twocolumn,letterpaper]{article}

\usepackage{iccv}
\usepackage{times}
\usepackage{epsfig}
\usepackage{graphicx}
\usepackage{amsmath}
\usepackage{amssymb}
\usepackage{comment}
\usepackage{graphics}
\usepackage{booktabs}
\usepackage{multirow}
\usepackage{enumitem}
\usepackage{bm}
\usepackage{adjustbox}
\usepackage[innercaption]{sidecap}
\usepackage{upquote}
\usepackage{textcomp}
\usepackage{pifont}
\usepackage{etoolbox}
\usepackage{caption}
\usepackage{subcaption}
\usepackage[utf8]{inputenc}
\usepackage{xspace}
\usepackage{enumitem}
\usepackage{lipsum}
\usepackage{tikz, pgfplots}
\usepackage[pagebackref=true,breaklinks=true,letterpaper=true,colorlinks,bookmarks=false]{hyperref}

\iccvfinalcopy % *** Uncomment this line for the final submission

\DeclareGraphicsExtensions{.jpeg,.png,.eps,.pdf}

\DeclareMathOperator*{\argmin}{arg\; min}
\newcommand{\der}[2]{\frac{\partial #1}{\partial #2}}

\newcommand{\mi}{\boldsymbol{-} \mathrel{\mkern -16mu} \boldsymbol{-}}

\ificcvfinal\fi
\begin{document}

%%%%%%%%% TITLE
%%%%%%%%% TITLE
\title{Model Decay in Long-Term Tracking}

\author{Efstratios Gavves\\
QUVA Lab, University of Amsterdam\\
{\tt\small e.gavves@uva.nl}
\and
Ran Tao$^1$\\
Kepler Vision Technologies\\
{\tt\small r.tao@keplervision.eu}
\and
Deepak K. Gupta\\
QUVA Lab, University of Amsterdam\\
{\tt\small d.k.gupta@uva.nl}
\and
Arnold W. M. Smeulders\\
QUVA Lab, University of Amsterdam\\
{\tt\small a.w.m.smeulders@uva.nl}
}

\maketitle

\footnotetext[1]{During this research, R. Tao was affiliated with QUVA Lab at University of Amsterdam.}
%\thispagestyle{empty}

%%%%%%%%% ABSTRACT
\begin{abstract}
Updating the tracker model with adverse bounding box predictions adds an unavoidable bias term to the learning. 
%In this paper, we highlight the negative impact that adverse bounding box predictions have to the tracker models, introducing an unavoidable bias term to the learning.
This bias term, which we refer to as model decay, offsets the learning and causes tracking drift. While its adverse affect might not be visible in short-term tracking, accumulation of this bias over a long-term can eventually lead to a permanent loss of the target. In this paper, we look at the problem of model bias from a mathematical perspective. Further, we briefly examine the effect of various sources of tracking error on model decay, using a correlation filter (ECO) and a Siamese (SINT) tracker. Based on observations and insights, we propose simple additions that help to reduce model decay in long-term tracking. The proposed tracker is evaluated on four long-term and one short term tracking benchmarks, demonstrating superior accuracy and robustness, even in 30 minute long videos.
%Model decay is stronger in long videos, simply because its gradual adverse effects accumulate over time.
%We examine what is the effect of various sources of tracking error on model decay, using a correlation filter and a siamese state-of-the-art trackers for a case study.
%Based on the empirical observations and insights, we propose simple additions that help with model decay and long-term tracking.
%The proposed tracker is evaluated on four long-term and one short term tracking benchmarks, demonstrating top accuracy and robustness, even in 30 minute long videos.
\end{abstract}

%%%%%%%%% BODY TEXT
\section{Introduction}
%%%%%%%%%%%%%
Fueled by the availability of standard datasets~\cite{kristan2015visual,wu2015object,smeulders2014visual}, tracking has made great progress over the last few years.
These datasets have mainly been designed to tackle challenges encountered in short-term tracking. For example, the average lengths of videos in ALOV~\cite{smeulders2014visual} and OTB~\cite{wu2015object} are only about 10 seconds and 20 seconds, respectively. The rationale behind designing these datasets was to select hard moments such as changes in illumination conditions, abrupt motion, clutter, large deformations, sudden occlusions, among others.

In general, performing well on the above datasets has been interpreted as overcoming the major challenges of tracking. However, in practice additional and different problems occur when the duration of tracking is longer, \emph{e.g.} half an hour.
When considering the practical applications of tracking, long durations are much more frequent than short videos, as in human interactions, sports, ego-documents and TV shows.

% However great their impact, and rightfully so, by these standards it was implicitly assumed that when most hard moments are solved, tracking of episodes in between will follow suit. In practice, however, it appears that in real-life scenarios, when tracking for half an hour, other elements then surviving the hard short episodes come also into play. And indeed, the half-an -hour time scale is important to most surveillance, man-machine interactions, sport games, ego-documents or TV show videos. 

To account for tracking challenges and appearance changes in short and long videos, state-of-art trackers must update their internal model frequently. However, too many model updates can eventually decay the inherent tracker model and the target might get lost. While the decay might not be noticeable in short-term videos, the accumulated effect is very prominent in long-term cases.
%Yet, one of the most crucial problem in tracking drifting, empirically recognized in~\cite{smeulders2014visual}.
%Tracking drifting can cause the model to gradually lose the object completely.
%In this work, we examine model decay, that is the worsening of the learned tracker model over long durations leading to drifting.
In this paper, we study model decay from a theoretical perspective with the goal of identifying underline reasons that cause it, and explore possible remedies to circumvent this problem.
% In this work we focus on a more theoretical examination of model decay in tracking and the reasons why this is caused.

% The model updates are generally based on temporally local information, and too many updates to the model can eventually deviate the tracker from its original target. For example, we  observed that even the state-of-art trackers such as ECO failed miserably on long videos.}}
% For long videos, the above cited difficult episodes may occur. But there is more. Tracking in long videos demonstrates challenges caused by the length of the video on which we focus here. 
% One of the most crucial problems is the tracking drift~\cite{smeulders2014visual}, which is caused by model decay.
% In this work we focus on model decay in tracking and the reasons why this is caused.

\begin{figure}[t!]   
\begin{center}
\includegraphics[width=0.4\textwidth, keepaspectratio]{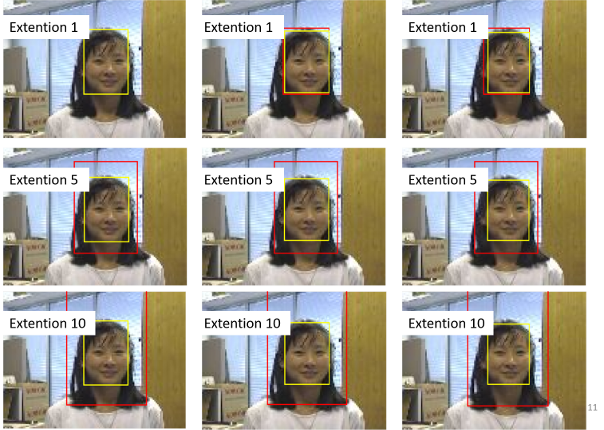}\label{fig:demo2}
\caption[]{Predictions from ECO \cite{Danelljan_2016_CVPR} on an artificially extended video created from OTB50 data (\textcolor{red}{red box}: tracker prediction, \textcolor{yellow}{yellow box}: ground truth prediction). \emph{Model decay} is prevalent here although the appearance variation remains intact. Due to heavy updating involved, model decay is noticeable from very early stages itself, even for clearly visible target objects moving slowly .}
\label{fig:teaser}
\vspace{-8mm}
\end{center}
\end{figure}

To demonstrate the severe effects of model decay on long videos, results from a simple numerical experiment are presented here. A sample video is chosen randomly from the OTB50 dataset, and extended artificially in the following manner: $[x_1, x_2, ..., x_{T-1}, x_T, x_T, x_{T-1}, ..., x_2, x_1, x_1, x_2, ..., x_T, ...].$. This way of extending ensures that any differences in tracking accuracy observed in the later parts of the long sequence will be caused solely due to increased video length. 

Fig. \ref{fig:teaser} shows the predictions obtained using ECO, one among the state-of-art trackers, for 3 different frames observed in 3 different repetitions of the original video. The 3 frames per repetition are chosen such that they lie close in the sequence and the appearance of the target does not change much among them. It can be seen that the tracker predictions become less and less accurate over time due to model decay caused by the gradual but erroneous, heavy updating. The tracker drift caused by model decay has been a long known phenomenon, but, in the context of short-term tracking, this issue has not been very relevant. To develop circumventing measures against it for long-term tracking, a theoretical understanding is needed that can provide a mathematical definition to the underlying mechanism.

%we take the state-of-the-art ECO tracker~\cite{Danelljan_2016_CVPR}, which relies on heavy model updating, and apply it on OTB videos, durations of which are extended artificially like $[x_1, x_2, ..., x_{T-1}, x_T, x_T, x_{T-1}, ..., x_2, x_1, x_1, x_2, ..., x_T, ...].$
%This artificially increases the length of the video without introducing additional visual appearance variations.
%In other words, any differences in tracking accuracy observed in the later parts of the long sequence will be caused solely due to increased video length.
%Fig. \ref{fig:teaser} shows a characteristic example, where model decay leads to significant accuracy drop although the video is seemingly quite easy.
%In this example the tracker predictions become less and less accurate over time due to model decay caused by the gradual but erroneous, heavy updating.

%Although tracker drift and model decay can be a serious challenge for long-term tracking, their underlying mechanisms are not entirely clear.
In this paper, we mathematically study the learning dynamics of general tracker models.
We show that in the presence of prediction errors, any model update adds always an unavoidable bias to the learning.
Second, based on numerical experiments, the correlation between model decay and various tracking challenges is studied on two different trackers: a correlation filter tracker and a siamese tracker.
Finally, based on the identified limitations of the trackers and gained insights, some simple modifications are proposed that allow a siamese tracker to deal better with model decay, thereby making it amenable to long-term tracking.

% As a proof of concept for the relevant of model decay in long-term tracking, we perform a simple experiment where we artificially extend the tracking video length of OTB~\cite{wu2015object} videos by repetition
% %
% \footnote{For a video with frames $x=[x_1, ..., x_T]$ we artificially increase the length by $x=[x_1, ..., x_T, x_T-1, ..., x_1, x_2, ...]$.}.
% %
% In Figure~\ref{fig:demo1} we plot the AUC tracking performance as a function of the number of repetitions for ECO~\cite{danelljan2017eco}, one of the leading trackers relying on model updates.
% As we observe similar behaviors in all videos, for clarity we focus on 10 random ones.
% Remarkably, ECO maintains its good performance for many of the videos even after several repetitions.
% Yet, in many other videos the model decay becomes apparent, even after the second repetition and even for clearly visible target objects that move slowly, like in Figure~\ref{fig:demo2}.

% We conclude that designing a long-term tracker model is a multi-faceted problem.
% While frame-to-frame accuracy is crucial, similar to short-term tracking, in long-term tracking one must also address model decay.

\section{Model Decay in Tracking}\label{sec:long-term}
% \subsection{Model Decay}

The most popular strategy for learning object trackers can be summarized in the following equations:
\begin{align}
    \phi_{t+1} & = \argmin_{\phi} \mathcal{L} (x_{1:t}, y_{1:t}) \\
    y_{t+1} & = f(x_{t+1}; \phi_{t+1}),
\end{align}
where $f$ is the $\phi$-parameterized tracker model that minimizes the tracker loss $\mathcal{L}$ over the dataset $D=[x_{1:t}, y_{1:t}]$ at the timestep $t+1$.
The dataset is composed of frames $x_{1:t}=[x_1, ..., x_t]$, and the tracker model $f$ returns as output the bounding box predictions $y_{1:t}=[y_1, ..., y_t]$. 
To reduce notation clutter, we use $f_{i, t}$ to refer to the output of the tracker model with parameters $\phi_t$ applied on frame $x_i$. 
The model parameters are updated by taking small steps towards the gradient direction of the loss surface, namely using gradient descent approach (or its variants),
\begin{align}
    \der{\phi}{t} & = -\eta \nabla_\phi \mathcal{L}_t \Rightarrow \\
    \phi_{t+1} & = \phi_t - \eta \nabla_\phi \mathcal{L}_t
    \label{eq:par-update}
\end{align}
Central to the tracking learning problem, therefore, is the gradient of the tracking loss with respect to the model parameters.
Extending on $\nabla_\phi \mathcal{L}$ and using the expectation over $t$ timesteps, $\mathbb{E}[\cdot]=\frac{1}{t}\sum_{i=1}^t [\cdot]$, we have
\begin{align}
    \nabla_\phi \mathcal{L}_t & = \nabla_\phi \mathbb{E} [(y_i-f_{i, t})^2] \label{eq:grad-loss-1} \\
    %& = \nabla_\phi \mathbb{E} [y_i^2+f_{i, t}^2 - 2y_if_{i, t}] \label{eq:grad-loss-2} \\
    & = 2\mathbb{E}[f_t\nabla_\phi f_t]-2\mathbb{E}[y_t\nabla_\phi f_t], \label{eq:grad-loss-3}.
\end{align}
To go from Eq. \eqref{eq:grad-loss-1} to \eqref{eq:grad-loss-3} we rely on that the bounding box coordinates $y_i$, predicted in previous frames, become dataset input variables with constant values.
Thus, they are independent of $\phi$ and $\mathbb{E}[\nabla_\phi y_i^2]=0$.
By substituting Eq.~\eqref{eq:grad-loss-3} in \eqref{eq:par-update}, the model parameters update can be described as
\begin{align}
\phi_{t+1} - \phi_t & = - 2\eta \Big[ \mathbb{E}[f_{i, t}\nabla_\phi f_{i, t}]-\mathbb{E}[y_i\nabla_\phi f_{i, t}] \Big]
\label{eq:model-update}
\end{align}

An interesting --but often overlooked reality-- is that while tracking is casted as a supervised learning problem, there is only one data sample in the learning dataset that is definitely correct.
This one and only correct sample is the pair $(x_1, y_1^*)$ defined by the user in the first frame, where $y_1^*$ represents the coordinates of the user specified bounding box describing the object.
While all other bounding boxes $y_i \enskip \forall \enskip i>1$ are used for re-training and fine-tuning the tracker, there is no guarantee that the bounding boxes $y_i$ are indeed correct or even good enough for learning.
In fact, had the predictions $y_i$ been good enough for re-training the tracker, the tracker would not need to be retrained.

Based on the argument presented above, it is reasonable to expect that the predictions, which also serve as future training samples for the re-training of the tracker, are noisy measurements of the true bounding box coordinates $y_i^*$. Assuming Gaussian noise with variance $\sigma_{i}^2$, we can state:
\begin{align}
    y_i & = y_i^* + \delta_{i}, \enskip \text{and} \enskip \delta_{i} \sim N(0, \sigma_{i}^2),
    \label{eq:box-annots}
\end{align}
With the substitution of Eq. \eqref{eq:box-annots} in \eqref{eq:model-update} and some rearrangement of terms, we have
\begin{align}
    \phi_{t+1} - \phi_t & = - 2\eta \Big[ \mathbb{E}[f_{i, t}\nabla_\phi f_{i, t}]-\mathbb{E}[(y_i^*+\delta_{i}) \nabla_\phi f_{i, t}] \Big] \\
    & =
    \underbrace{
    - 2\eta\mathbb{E} [(f_{i, t}-y_i^*) \cdot \nabla_\phi f_{i, t}]
    }_{\text{Perfect parameter update}} + 
    \underbrace{
    2\eta\mathbb{E} [\delta_{i} \cdot \nabla_\phi f_{i, t}]
    }_{\text{Parameter bias}}
    \label{eq:perfect-bias}
\end{align}
It is easy to recognize the two components in the parameter update of the tracker.
The first term in Eq.~\eqref{eq:perfect-bias} corresponds to the perfect model update component, as it corrects the error made by the model prediction $f_{i, t}$ as compared to the perfect box $y^*_i$.
The second term corresponds to the parameter bias component, as this term depends directly on the error made by past predictions $\delta_{i}$.
If $\delta_{i}=0$, then there would be no error, and the parameter updates would also be perfect.

\paragraph{Model dynamics.}
Having computed the effect of the past errors on the parameter updates of the tracker, we can next examine the effect on the model dynamics $\der{f}{t}$ over time.
Specifically, after updating the parameters, the relation between the past $f_{i, t}$ and the next model $f_{i, t+1}$ is
\begin{align}
\der{f}{t} \propto f_{i, t+1}-f_{i, t} & = \der{f}{\phi}\der{\phi}{t} \Rightarrow \\
f_{i, t+1} & = f_{i, t} + \der{f}{\phi}\der{\phi}{t}
\label{eq:chain}
\end{align}
As $\der{\phi}{t} \propto \phi_{t+1} - \phi_t$, combining eq.~\eqref{eq:chain} and ~\eqref{eq:perfect-bias} we have that
\begin{align}
    \label{eq:dynamics}
    f_{i, t+1} = f_{i, t} 
    &
    \underbrace{
    -2\eta\mathbb{E} [(f_{i, t}-y_i^*) \cdot \|\nabla_\phi f_{i, t}\|^2]
    }_{\text{Perfect model update}} \\ \nonumber
    &
    \underbrace{
    + 2\eta\mathbb{E} [\delta_{i, t} \cdot \|\nabla_\phi f_{i, t}\|^2]
    }_{\text{Model decay}}
\end{align}
From Eq.~\eqref{eq:dynamics}, we make the following observation.
Due to the continuous updates, the tracker model offshoots its predictions by a quantity that is linearly proportional to past errors. We refer to this quantity as model decay.
%By taking the expectation of $\delta_{i, t} \cdot \|\nabla_\phi f_{i, t}\|^2$ over time, we can easily monitor --as an oracle on a held-out set, since the perfect predictions are not available-- the bias added to the tracker model and examine the generalization of the learned tracker.
%It is important to note that not all tracker updates are adverse: some updates will be beneficial while others will be not. For long-term tracking, the target can significantly change its appearance, and in such scenarios, the tracker model needs to be adjusted for the new look. Thus, updates are definitely needed. However, it is important that updates are performed cautiously. 

\paragraph{Long-term tracking \& model decay.}
As the model dynamics in eq.~\eqref{eq:dynamics} are recursive, it is implied that the bias term accumulates and, in fact, worsens over time.
For as long as the cumulative model decay is small enough, usually in the early iterations, the model dynamics is sufficiently accurate.
The early errors $\delta_{i,}$ are small not only because the tracker is still accurate, but also because the number of summands $t$ is small.
This is the reason why model decay is not a problem, and often goes even unnoticed, in short  videos up to 10-20 sec.
However, in longer videos where the number of summand $t$ grows and the $\delta_{i, t}$ errors grow as well, the cumulative model decay become noticeable.
The tracking paradox is more and more relevant when considering longer videos.

% \vspace{-1em}

\paragraph{Dynamics of correlation filter and siamese trackers.}
Based on the model dynamics described, we further analyze two trackers chosen from correlation filter (ECO \cite{Danelljan_2016_CVPR}) and siamese (SINT \cite{tao2016sint}) categories.

Regarding the initial model, correlation filter trackers rely on training using the first frame sample (and variants) for training.
They also perform many model updates, even as frequently as every frame.
These trackers are known to perform well on short-term videos. However, in longer videos the cumulative model bias should be expected to grow very large, and this will have a significant impact on the model dynamics, thereby causing tracker drift.

In contrast, siamese trackers rely for their initial model on a similarity function trained externally on large datasets of matching versus non-matching visual pairs.
Since siamese trackers can perform well even with no updates~\cite{tao2016sint} their cumulative model decay is 0.
The reason is that although prediction errors $\delta_{i}$ occur, the model us never updated, so $\nabla_\phi f_{i, t}=0$.
This means, however, that the tracker relies for its total model dynamics on the initial model only. All in all, in long videos siamese trackers are expected to show little or no model decay. However, their performance is sensitive in short videos and is only as good as their initial matching model, as empirically observed in~\cite{valmadre2018oxuva}.

\paragraph{An applied perspective.} The consequence of the theoretical framework provided above is that now the aptitude of updating can be measured, and the presented mathematical study lays the ground work for designing remedies to reduce the model bias. While a clear solution to avoid model decay is not yet known, the obtained insights provide us a simple yet effective approach to reduce model bias to a certain extent.

From Eq. \ref{eq:perfect-bias}, it can be seen that an accurate estimation of model bias would require computing the term $\mathbb{E}\left[\delta_i\nabla_{\phi}f_{i,t}\right]$. However, the state parameters $\phi_{t+1}$ are not known beforehand, and redetecting the target in the earlier frames for every model update step (to obtain $f_{i,t}$) will be computationally very expensive. While a feasible approximation to the bias term in Eq. \ref{eq:perfect-bias} is left for future research, we use a very simplified version of the bias term (by controlling $\delta$). To demonstrate the effectiveness of the process, we present LT-SINT (a long-term variant of the siamese tracker SINT). Note that although SINT has been chosen to demonstrate the concept, the underlying idea is general, and can be combined with any existing tracker without much modification.

Under the scenarios of abrupt motion, SINT can lose the target and the false positives encountered thereafter increase $\delta$ quickly, adding remarkable residual to the model bias. To reduce this, LT-SINT employs a heuristic scheme where global search is performed every $T$ frames. This reduces the chances of a target being lost for more than $T$ frames at once. Compared to the local search, the global search results are more reliable and tracker model can be updated at these instances. However, occasional large appearance variations of the target coinciding with the global search step can lead to an even severe false positive getting identified during this step. This is definitely not desired since using this prediction to update the model would lead to large values of $\delta$ as well as adversely affect all terms defined by $f{i,t+1}$. To circumvent this issue, a motion model (referred to as \emph{decay recognition network}) is introduced, which ensures that unreliable estimates during global search are prevented from contributing towards model updates. More details related to LTSINT are presented in Section \ref{sec_ltsint}.

\section{Tracking Challenges \& Model Decay}
\subsection{Long OTB}

In~\cite{wu2013online, smeulders2014visual} the different tracking challenges have been systematically studied.
As we would like to study the contribution of these sources of errors in model decay, we extend the videos from OTB-50~\cite{wu2013online} dataset as follows.
For a video with frames $x=[x_1, ..., x_T]$ we artificially increase its length by $x=[x_1, ..., x_T, x_T-1, ..., x_1, x_2, ...]$, where a forward and reverse pass, $x_1, ..., x_T, x_{T-1}, ..., x_1$ counts as a single repetition.
As the metadata annotations for the different tracking challenges are video-wide, the same annotations apply also for the extended videos.
We refer to this dataset as the \emph{Long OTB} and we propose to use it to monitor model decay in visual object trackers.

\subsection{Effect of Tracking Challenges on Model Decay}

%---------Figure: sources of error------------------------------------------------------
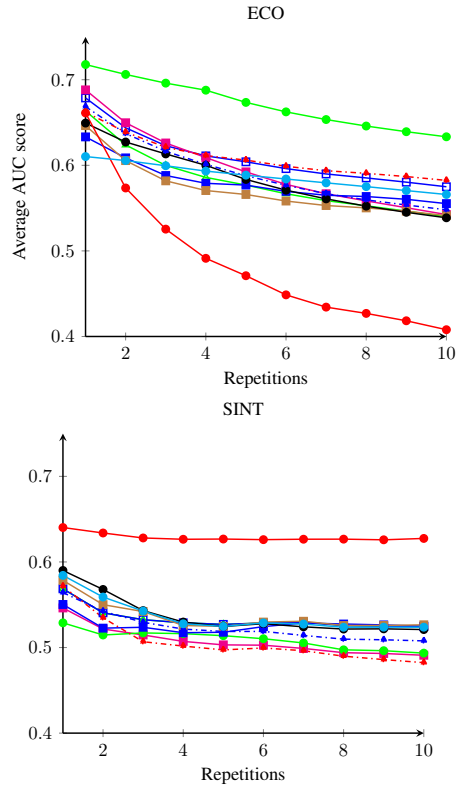
\begin{figure} 
	\begin{center}
	   	\begin{tikzpicture}[scale = 0.7]
		\begin{axis}[%
		thick, 
		axis x line=bottom,
		axis y line=left,
		legend style={at={(0.5,1.7)},anchor=north},
		%legend pos = outer north east,
		xlabel = {Repetitions},
		ylabel = {Average AUC score},
		ymax = 0.75,
		ymin=0.4,
		title={ECO}
		%ylabel style = {rotate = -90},
		%ymode = log
		]
		%\addlegendimage{empty legend}
e		\addplot[mark=triangle, mark options={scale=1}, solid, green,  thick] table[x=rep, y = IV] {data_files/eco_auc.dat}; \label{line:iv}
		\addplot[mark=square, mark options={scale=1}, solid, blue,  thick] table[x=rep,y = SV] {data_files/eco_auc.dat}; \label{line:sv}
		\addplot[mark=square*, mark options={scale=1}, magenta, thick] table[x=rep, y = OCC] {data_files/eco_auc.dat}; \label{line:occ}
		\addplot[mark=*, mark options={scale=1}, solid, green, thick] table[x=rep, y = DEF] {data_files/eco_auc.dat};\label{line:def}
		\addplot[mark=square*, mark options={scale=1}, solid, blue, thick] table[x=rep,y = MB] {data_files/eco_auc.dat};\label{line:mb}
		\addplot[mark=square*, mark options={scale=1}, solid, brown, thick] table[x=rep, y = FM] {data_files/eco_auc.dat};\label{line:fm}
		\addplot[mark=*, mark options={scale=1}, solid, black, thick] table[x=rep, y = IPR] {data_files/eco_auc.dat};\label{line:ipr}
		\addplot[mark=triangle*, mark options={scale=1}, dashdotted, blue, thick] table[x=rep, y = OPR] {data_files/eco_auc.dat};\label{line:opr}
		\addplot[mark=*, mark options={scale=1}, solid, red, thick] table[x=rep, y = OV] {data_files/eco_auc.dat};\label{line:ov}
		\addplot[mark=triangle*, mark options={scale=1}, dashdotted, red,  thick] table[x=rep, y = BC] {data_files/eco_auc.dat};\label{line:bc}
		\addplot[mark=*, mark options={scale=1}, solid, cyan, thick] table[x=rep, y = LR] {data_files/eco_auc.dat};\label{line:lr}
% 		\addlegendentry{Illumination Variation}
% 		\addlegendentry{Scale Variation}
% 		\addlegendentry{Occlusion}
% 		\addlegendentry{Deformation}
% 		\addlegendentry{Motion Blur}
% 		\addlegendentry{Fast Motion}
% 		\addlegendentry{In-Plane Rotation}
% 		\addlegendentry{Out-of-Plane Rotation}
% 		\addlegendentry{Out-of View}
% 		\addlegendentry{Background Clutters}
% 		\addlegendentry{Low Resolution}
		\end{axis}
		\end{tikzpicture}
% 		\vspace{-1em}
		\begin{tikzpicture}[scale = 0.7]
		\begin{axis}[%
		thick, 
		axis x line=bottom,
		axis y line=left,
		legend style={at={(0.5,1.7)},anchor=north},
		%legend pos = outer north east,
		xlabel = {Repetitions},
		ymax = 0.75,
		ymin=0.4,
		title={SINT}
		%ylabel style = {rotate = -90},
		%ymode = log
		]
		%\addlegendimage{empty legend}
e		\addplot[mark=triangle, mark options={scale=1}, solid, green,  thick] table[x=rep, y = IV] {data_files/sint_auc.dat};
		\addplot[mark=square, mark options={scale=1}, solid, blue,  thick] table[x=rep,y = SV] {data_files/sint_auc.dat};		
		\addplot[mark=square*, mark options={scale=1}, magenta, thick] table[x=rep, y = OCC] {data_files/sint_auc.dat};
		\addplot[mark=*, mark options={scale=1}, solid, green, thick] table[x=rep, y = DEF] {data_files/sint_auc.dat};
		\addplot[mark=square*, mark options={scale=1}, solid, blue, thick] table[x=rep,y = MB] {data_files/sint_auc.dat};
		\addplot[mark=square*, mark options={scale=1}, solid, brown, thick] table[x=rep, y = FM] {data_files/sint_auc.dat};
		\addplot[mark=*, mark options={scale=1}, solid, black, thick] table[x=rep, y = IPR] {data_files/sint_auc.dat};
		\addplot[mark=triangle*, mark options={scale=1}, dashdotted, blue, thick] table[x=rep, y = OPR] {data_files/sint_auc.dat};
		\addplot[mark=*, mark options={scale=1}, solid, red, thick] table[x=rep, y = OV] {data_files/sint_auc.dat};
		\addplot[mark=triangle*, mark options={scale=1}, dashdotted, red,  thick] table[x=rep, y = BC] {data_files/sint_auc.dat};
		\addplot[mark=*, mark options={scale=1}, solid, cyan, thick] table[x=rep, y = LR] {data_files/sint_auc.dat};
		\end{axis}
		\end{tikzpicture}
		\label{fig:difficulties-decay-sint}
	\end{center}
	\caption{The effect of the 11 sources of tracking difficulty in Long OTB-50 videos. The legend goes as follows: illumination variation-IV (\ref{line:iv}), scale variation-SV (\ref{line:sv}), occlusion-OCC (\ref{line:occ}), deformation-DEF (\ref{line:def}), motion blur-MB (\ref{line:mb}), fast motion-FM (\ref{line:fm}), in-plane rotation-IPR (\ref{line:ipr}), out-of-plane rotation-OPR (\ref{line:opr}), out-of-view-OV (\ref{line:ov}), background clutter-BC (\ref{line:bc}), low resolution-LR (\ref{line:lr}).}
	\label{fig:difficulties-decay}
\end{figure}

%---------------------------------------------------------------------------------------
Next, how model decay is influenced by the various tracking challenges documented in tracking surveys~\cite{wu2013online, smeulders2014visual}.
Specifically, we focus on the tracking challenges reported in OTB-50~\cite{wu2013online}: illumination variation, scale variation, occlusion, deformation, motion blur, fast motion, in-plane rotation, out-of-plane rotation, out-of-view, background clutters, low resolution.

In long videos, some of these tracking challenges may become extreme, leading to stronger pressure towards model decay. 
We identify three additional, extreme challenges: sampling drift, target disappearance and re-appearance, and pixel-level errors.
Sampling drift relates to the fast motion and scale variation challenges.
It is caused by imperfect local scale-space search~\cite{hare2011struck,HenriquesC0B15,DanelljanBMVC2014,danelljan2017eco,tao2016sint,bertinetto2016fully,nam2016learning}, often due to the target exceeding a user-defined maximum speed.
Target disappearance and reappearance is an extreme version of the out-of-view and occlusion challenges, where the target object leaves the frame completely for an unknown length of time.
As a consequence, there is no guarantee for the location of the re-entrance, compromising the assumption of motion continuity.
Pixel-level errors refer to the label errors.
They happen especially during generation of the ``pseudo-positives'' based on \emph{ad hoc} rules like maximum overlap over time, which can be satisfied even with wrong detections. 
Although a $<0.7$ IoU threshold is good enough for evaluating tracker accuracy, the added background pixels start counting as positive appearances, compounding slowly but consistently and leading to serious model decay.
Pixel-level errors is a challenge relates to learning and optimization of the tracker and connects to all tracking challenges.

We plot the average AUC curves for the eleven annotated tracking challenges in Fig.~\ref{fig:difficulties-decay} (up) for ECO~\cite{Danelljan_2016_CVPR} and Fig.~\ref{fig:difficulties-decay} (bottom) for SINT.
Although we have no annotations for sampling drift, target disappearance and re-appearance, and pixel-level errors, they connect to the eleven annotated tracking challenges and we can thus derive insights indirectly.

There are several interesting observations.
First, ECO --relying on heavy model updating-- pays a particularly high price when the target is occluded ($\sim 10\%$ drop) or goes out-of-view ($\sim 25\%$ drop) throughout the video.
The other tracking challenges also contribute to model decay in the order of $~\sim 3-8 \%$.
The only tracking challenges that appears to not contribute much to model decay is low resolution, probably because of the multi-scale search of ECO.

In contrast, SINT seems to be less affected by model decay. Since there is no model update mechanism in SINT, one would expect observing no model bias with it. However, due to the inherent local search strategy SINT can occasionally lose the target and this extent of losing can var across repetitions. Due to this reason, a slight drop in AUC score can be observed over several repetitions.
The biggest problem appears to be in-plane and out-of-plane rotation\footnote{Without any provisions standard convolutional neural networks are not rotation invariant, so in-plane rotations generate different features}. ($\sim 5\%$ drop) and background clutter ($\sim 7\%$ drop), while for the rest of the challenges the drop is in the order of $2-5\%$.
That said, it is also clear that SINT scores considerably lower that that of ECO.

While the siamese tracker seems to perform well, it is also on a disadvantage as the first frame cannot alone describe the full appearance of the target object.
Since model decay impacts long-term tracking significantly, siamese trackers with careful model updating is a reasonable forward direction. Last, we emphasize this analysis contemplates the best case scenario of having the same video frames repeated again and again. In reality, long videos change their appearance significantly and the above tracking challenges are expected to be even harsher.

\section{Minimizing Model Decay, Simply}

\subsection{Predicting Model Decay}

While eq.~\eqref{eq:dynamics} explains the effect of the model decay on the dynamics of the model, it is only a theoretical tool.
The reason is that it requires the coordinates of the optimal bounding box around the target, which is what we are looking for in the first place.
by re-arranging the terms of eq.~\eqref{eq:dynamics} and setting $\omega_t=(1 - \frac{\delta_{i, t}}{f_{i, t}-y_i^*})$, however, we have that 
\begin{align}
    f_{i, t+1} = f_{i, t} - 2\eta \cdot \mathbb{E} [\omega_t \cdot (f_{i, t}-y_i^*) \|\nabla_\phi f_{i, t}\|^2]
    \label{eq:decay}
\end{align}
One way to interpret eq.~\eqref{eq:decay} is that $\omega_t \in \{0, 1\}$ is a binary weight variable, which allows for model updates for only a fraction of the frames when $\omega_t=1$.
A logical step then is to measure $\omega_t$ by an independent model, \eg, a second neural network, trained to recognize if the next frame is resembles a perfect or biased model update.
We will refer to this as the \emph{decay recognition network}.

To make sure any bias in $\nabla_\phi f_{i, t}$ has a minimal effect on $\omega_t$, the decay recognition network must share as little correlation as possible with the tracker model.
This means the decay recognition network has as little correlation as possible to the tracker model should not share any parameters with the tracker model.
Also, it should not receive overly correlated input patches. 

The latter is clearly a hard requirement, as by definition tracker predictions usually relate to each other in nearby frames.
A solution is to rely on siamese trackers~\cite{tao2016sint}, whose tracking predictions are independent over time and suffer less from model decay already.
Although the focus of this work is not to present a fully-fledged tracker that solves model decay, we, therefore, present two simple modifications on SINT~\cite{tao2016sint}.

\subsection{A Simple Long-term Tracker}
\label{sec_ltsint}
%-------------Figure on AUC comparisons----------------------------------------------------
\begin{figure*} 
	\centering
	\begin{subfigure}{0.33\textwidth}
	\begin{center}
		\begin{tikzpicture}[scale = 0.65]
		\begin{axis}[%
		ymax = 1,
		ymin=0,
		thick, 
		axis x line=bottom,
		axis y line=left,
		legend style={at={(0.5,1.7)},anchor=north},
		%legend pos = outer north east,
		xlabel = {Repetition},
		ylabel = {AUC score}
		%ylabel style = {rotate = -90},
		%ymode = log
		]
		%\addlegendimage{empty legend}
		\addplot[mark=triangle*, mark options={scale=1}, solid, red,  thick] table[x=rep, y = auc] {data_files/ECO/girl.dat}; \label{line:eco-girl}
		\addplot[mark=square*, mark options={scale=1}, solid, red,  thick] table[x=rep, y = auc] {data_files/ECO/Freeman4.dat};\label{line:eco-freeman4}
		\addplot[mark=diamond*, mark options={scale=1}, solid, red,  thick] table[x=rep, y = auc] {data_files/ECO/Lemming.dat};\label{line:eco-lemming}
		\addplot[mark=otimes*, mark options={scale=1}, solid, red,  thick] table[x=rep, y = auc] {data_files/ECO/Suv.dat};\label{line:eco-suv}
		\addplot[mark=triangle, mark options={scale=1}, solid, blue,  thick] table[x=rep, y = auc] {data_files/ECO/Car4.dat};\label{line:eco-car4}
		\addplot[mark=square, mark options={scale=1}, solid, blue,  thick] table[x=rep, y = auc] {data_files/ECO/Couple.dat};\label{line:eco-couple}
		\addplot[mark=diamond, mark options={scale=1}, solid, blue,  thick] table[x=rep, y = auc] {data_files/ECO/Skiing.dat};\label{line:eco-football1}
		\addplot[mark=otimes*, mark options={scale=1}, solid, blue,  thick] table[x=rep, y = auc] {data_files/ECO/Freeman1.dat};\label{line:eco-freeman1}
		%\addlegendentry{$r_{\text{min}}$}
		\end{axis}
		\end{tikzpicture}
	\end{center}\vspace{-1.5em}	
	\caption{ECO-deep}
	\end{subfigure}
	\begin{subfigure}{0.33\textwidth}
	\begin{center}
		\begin{tikzpicture}[scale = 0.65]
		\begin{axis}[%
		ymax = 1,
		thick, 
		axis x line=bottom,
		axis y line=left,
		legend style={at={(0.5,1.7)},anchor=north},
		%legend pos = outer north east,
		xlabel = {Repetition},
		ylabel = {AUC score}
		%ylabel style = {rotate = -90},
		%ymode = log
		]
		%\addlegendimage{empty legend}
		\addplot[mark=triangle*, mark options={scale=1}, solid, red,  thick] table[x=rep, y = auc] {data_files/SINT/girl.dat};\label{line:lteco-girl}
		\addplot[mark=square*, mark options={scale=1}, solid, red,  thick] table[x=rep, y = auc] {data_files/SINT/Freeman4.dat};\label{line:lteco-bolt}
		\addplot[mark=diamond*, mark options={scale=1}, solid, red,  thick] table[x=rep, y = auc] {data_files/SINT/Lemming.dat};
		\addplot[mark=otimes*, mark options={scale=1}, solid, red,  thick] table[x=rep, y = auc] {data_files/SINT/Suv.dat};
		\addplot[mark=triangle, mark options={scale=1}, solid, blue,  thick] table[x=rep, y = auc] {data_files/SINT/Car4.dat};
		\addplot[mark=square, mark options={scale=1}, solid, blue,  thick] table[x=rep, y = auc] {data_files/SINT/Couple.dat};
		\addplot[mark=diamond, mark options={scale=1}, solid, blue,  thick] table[x=rep, y = auc] {data_files/SINT/Skiing.dat};
		\addplot[mark=otimes*, mark options={scale=1}, solid, blue,  thick] table[x=rep, y = auc] {data_files/SINT/Freeman1.dat};
		%\addlegendentry{$r_{\text{min}}$}
		\end{axis}
		\end{tikzpicture}
	\end{center}\vspace{-1.5em}	
	\caption{SINT}
	\end{subfigure}
	\begin{subfigure}{0.33\textwidth}
	\begin{center}
		\begin{tikzpicture}[scale = 0.65]
		\begin{axis}[%
		ymax = 1,
		ymin=0,
		thick, 
		axis x line=bottom,
		axis y line=left,
		legend style={at={(0.5,1.7)},anchor=north},
		%legend pos = outer north east,
		xlabel = {Repetition},
		ylabel = {AUC score}
		%ylabel style = {rotate = -90},
		%ymode = log
		]
		%\addlegendimage{empty legend}
		\addplot[mark=triangle*, mark options={scale=1}, solid, red,  thick] table[x=rep, y = auc] {data_files/LTSINT/girl.dat}; \label{line:ltsint-girl}
		\addplot[mark=square*, mark options={scale=1}, solid, red,  thick] table[x=rep, y = auc] {data_files/LTSINT/Freeman4.dat}; \label{line:ltsint-freeman4}
		\addplot[mark=diamond*, mark options={scale=1}, solid, red,  thick] table[x=rep, y = auc] {data_files/LTSINT/Lemming.dat}; \label{line:ltsint-lemming}
		\addplot[mark=otimes*, mark options={scale=1}, solid, red,  thick] table[x=rep, y = auc] {data_files/LTSINT/Suv.dat}; \label{line:ltsint-suv}
		\addplot[mark=triangle, mark options={scale=1}, solid, blue,  thick] table[x=rep, y = auc] {data_files/LTSINT/Car4.dat}; \label{line:ltsint-car4}
		\addplot[mark=square, mark options={scale=1}, solid, blue,  thick] table[x=rep, y = auc] {data_files/LTSINT/Couple.dat}; \label{line:ltsint-couple}
		\addplot[mark=diamond, mark options={scale=1}, solid, blue,  thick] table[x=rep, y = auc] {data_files/LTSINT/Skiing.dat}; \label{line:ltsint-skiing}
		\addplot[mark=otimes*, mark options={scale=1}, solid, blue,  thick] table[x=rep, y = auc] {data_files/LTSINT/Freeman1.dat}; \label{line:ltsint-freeman1}
		%\addlegendentry{$r_{\text{min}}$}
		\end{axis}
		\end{tikzpicture}
	\end{center}\vspace{-1.5em}		
	\caption{LT-SINT}
	\end{subfigure}
	\caption{Running ECO, SINT, and LT-SINT on eight Long OTB videos, four videos with significant model decay (\textcolor{red}{$\mi$}) and four videos with small model decay (\textcolor{blue}{$\mi$}), for ECO and SINT. The videos are: Girl: SV, OCC, IPR, OPR (\ref{line:ltsint-girl}), Freeman4: SV, OCC, IPR, OPR (\ref{line:ltsint-freeman4}), Lemming: IV, SV, OCC, FM, OPR, OV (\ref{line:ltsint-lemming}), SUV: OCC, IPR, OV (\ref{line:ltsint-suv}), Car4: IV, SV (\ref{line:ltsint-car4}), Couple: SV, DEF, FM, OPR, BC (\ref{line:ltsint-couple}), Skiing: IV, SV, DEF, IPR, OPR (\ref{line:ltsint-skiing}), Freeman1: SV, IPR, OPR (\ref{line:ltsint-freeman1}).}
	\label{fig:hard-easy-cases}
\end{figure*}
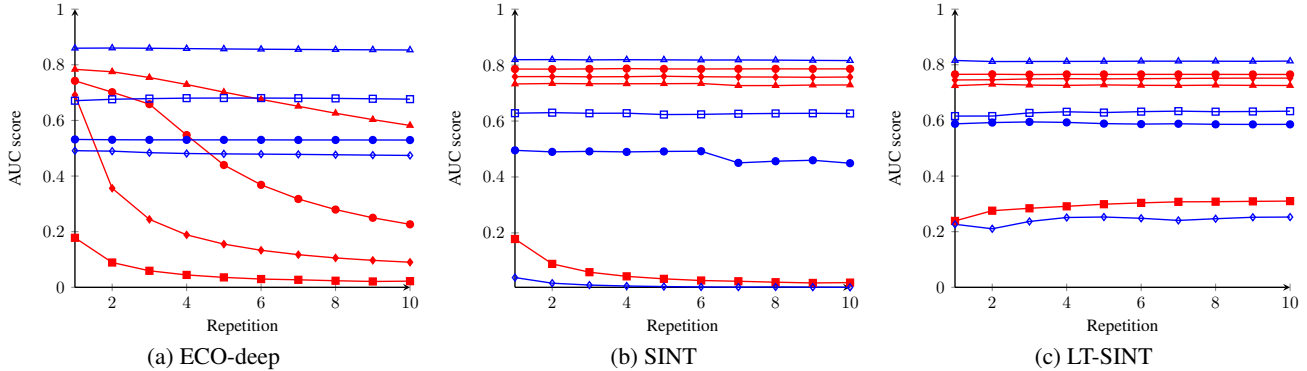
%------------------------------------------------------------------------------------------
SINT~\cite{tao2016sint} performs no model updates and local search.
So, the tracker follows the target only for as long as its appearance does not deviate too much from the first frame and the local search space contains the target. To account for significant appearance changes and sampling drift, two improvements are applied over SINT to design LTSINT. 

First, we introduce a decay recognition network (DRN) that regulates whether the next model update will contribute bias or not, similar to self-awareness~\cite{MaYZY15}. With the DRN module, we expect to obtain a qualitative approximation of $\omega_t$. While there is no fixed choice for it, our DRN conditions on the similarity map in the current frame as well as the previous $K-1$ frames. An LSTM-based binary classifier is used, and similarity maps from the history are incorporated to capture the temporal dynamics of the similarity distributions. Since the DRN operates on the predictions from the previous frames, the risk of being getting infected by model bias still pertains. To make sure no such additional bias is introduced, we simply train the LSTM using similarity feature maps from the original siamese similarity function only. Specifically, training is performed using tracks $z$ from the standard siamese tracker~\cite{tao2016sint} on a held out set deemed positive (or negative) when having an $>0.5$ (or $<0.5$) IoU threshold. At test time, when the LSTM scores above a highly conservative threshold, the siamese similarity function is updated.

The second change is to perform hybrid search to locate the most likely next position of the target. This involves performing local search, as in \cite{tao2016sint}, as well as searching globally after every $T$ frames. The addition of global search reduces the extent to which a target can be lost in one instance. Moreover, since it is performed only sparsely, the increase in computational time is limited. Global search involves a hierarchical search strategy of three levels. First, the $N (=10)$ mos probable locations are identified. Around each of the $N$ locations, search is then performed at $M$ different scales during the second stage. For better localization in spatial and scale spaces, the third stage involves analysis over $L$ finer scales around the $M\cdot N$ selected locations. The candidate with the highest value on the similarity map is then chosen. Local search is similar to the third stage of global search, and target is searched around the prediction obtained in the previous frame.

All said, we emphasize that no matter how careful we are with our model updates, the risk of model decay still exists. In practice, with the suggested modifications we did not witness significant model decay for videos even up to 30 minutes long.

\paragraph{Model Decay on LT-SINT}

We run ECO, SINT, LT-SINT on eight Long OTB videos, four videos with significant model decay with red and four videos with small model decay with blue, and show results in Fig.~\ref{fig:hard-easy-cases}.
We observe that LT-SINT maintains top accuracy and is impervious to model decay, maintaining tracking throughout the whole duration of the video.
In fact for some videos where ECO takes a toll due to model decay, like Freeman4, LT-SINT even improves.
As LT-SINT appears to cope better with model decay, we focus on LT-SINT for the final benchmarking.

\section{Related Work}

\paragraph{Long-term tracking.}
To our knowledge, only limited works exist that focus on long-term tracking. A seminal work on long-term tracking is the TLD~\cite{kalal2012tracking} tracker. It is a multi-component method that combines an optical flow tracker with a detector, and the detector part is slowly updated. However, TLD is very sensitive to tracking circumstances. SPL~\cite{supancic2013self} follows the TLD paradigm for long-term tracking, and employs a SVM-based detector which is updated using the frames that produce the lowest SVM-objective for the training set. The repeated evaluation of the SVM objective, however, has a noticeable computational footprint.

LTCT~\cite{MaYZY15} is a correlation filter tracker paying particular attention to long-term tracking, by integrating an online detector to re-detect the target in case of tracking failures. Furthermore, in the recent long-term challenge by VOT several new trackers were proposed~\cite{Kristan2018a}, grouped into four families: short-trackers that do not implement re-detection nor model occlusion, short-term trackers with conservative updating, pseudo  long-term trackers that do not return a box when the target is not visible, and re-detecting long-term trackers.

Recently, siamese paradigm-based trackers have been proposed which are shown to deliver state-of-art performance on long-term videos. Some such trackers include MBMD \cite{Kristan2018a}, DaSiamLT \cite{Kristan2018a}, SiamRPN++ \cite{Li2019siamrpnpp} \emph{etc.}. On top of the Siamese framework, these trackers use a SSD-MobielNet architecture, a region proposal framework and deep ResNet50 architectures, respectively, which help them to generate very accurate bounding box predictions.

As until recently few works only focused on long-term tracking, only few long-term benchmark datasets were available also.
The most popular dataset to this year was UAV20L with 20 long videos captured from low-altitude unmanned aerial vehicles.
Interestingly, in last few months several new benchmarks~\cite{valmadre2018oxuva, moudgil2017long, mueller2016benchmark, Kristan2018a, MilanL0RS16, TrackingNet} were introduced, several of whom focus on long-term tracking.

In this work we analytically analyze the impact of model decay, among others its model decay which leads to drift~\cite{smeulders2014visual}.
We propose a tracker that relies on Siamese instance search with rare and self-aware, and hence robust, model updates.
Further, we revisit the importance of full image search, originally proposed in~\cite{kalal2012tracking}, for long-term tracking. 

\paragraph{Short-term tracking.}
Short-term tracking has been one of the oldest and most popular fields in computer vision.
Short-term trackers can be grouped in three categories: those that perform tracking by detection, tracking by correlation filters and tracking by similarity comparisons.

MDNet~\cite{nam2016learning} is a recent successful tracking-by-detection tracker using deep learning.
It employs a deep classification network as the detector, while the last layer is specialized for each video. MDNet employs a risky update strategy and a local search scheme, thus being sensitive tracking and sampling drift. 
EBT~\cite{Zhu_2016_CVPR} goes beyond local search by generating instance-specific object proposals over the whole frame, using an SVM learned online.
The online learning of EBT runs also the risk of model decay even in the stage of proposal generation. 

Very good performance is delivered by trackers based on discriminative correlation filters (DCF).
Since the MOSSE tracker~\cite{BolmeBDL10}, many variants have been proposed.
~\cite{Danelljan_2014_CVPR} uses multi-dimensional features.
~\cite{HenriquesC0B15} proposes kernelized correlation filters.
~\cite{DanelljanBMVC2014} incorporate robust scale estimation, while~\cite{Danelljan2015,kiani2015correlation} address the boundary effects caused by the circular shift.
~\cite{danelljan2016beyond} learns the filters in the continuous spatial domain and ~\cite{danelljan2017eco} further improves~\cite{danelljan2016beyond} by improving on the model over-fitting.
And,~\cite{ma2015hierarchical,qi2016hedged,danelljan2016beyond,danelljan2017eco} integrate convolutional features with the DCF framework.
DCF trackers have shown great performance on tracking benchmarks of short videos~\cite{kristan2015visual,wu2015object}. However, by relying on frequent, risky update schemes and local search, these trackers are susceptible to model decay in long videos.

Recently, a new and promising family is Siamese trackers.
The first Siamese trackers~\cite{tao2016sint, bertinetto2016fully} relied on a tracking by similarity comparison strategy.
They simply search for the candidate most similar to the original image patch of the target given in the starting frame, using a run-time fixed but learned \textit{a priori} deep Siamese similarity function.
Due to their no-updating nature, Siamese trackers are robust against several sources of error that cause model decay. Improved versions of siamese trackers include regional proposal frameworks (\emph{e.g.} SiamRPN \cite{Li_2018_CVPR}, DaSiamRPN \cite{Zhu2018dasiamrpn}) and deeper backbones such as ResNet50 (SiamRPN++ \cite{Li2019siamrpnpp}) which provide more accurate predictions and improve the tracking performance.

\section{Experiments}
\subsection{Implementation Details}

For LT-SINT we adopt most implementation details from SINT.
We rely on an Imagenet pretrained VGG-16 network~\cite{simonyan2015very} up to the \texttt{relu4\_3} layer, consisting of 10 convolution layers, 3 2-by-2 max pooling layers, and ReLU~\cite{nair2010rectified} nonlinearities. The first stage of global search identifies 10 locations and in the second stage, the search is performed over 11 scales: $2^{\{-0.4:0.08:0.4\}}$. In these two stages, the query patch of the initial target is resized to $32 \times 32$ pixels. For the third stage as well as the local search strategy, search is performed over 5 scales close to the previously estimated one: $\{0.9509,0.9751,1,1.0255,1.0517\}$~\cite{bertinetto2016fully}. The input resolution for these cases is set to $64\times 64$ pixels.

We train the LSTM decay recognition network offline on tracks returned by SINT with global search on ALOV~\cite{smeulders2014visual}. In the LSTM network, the similarity maps are first encoded by a small convnet and the whole sequence is summarized by a two-layer LSTM network following by two fully-connected layers. The convnets comprise two $3\times3$ convolutional layers with strides of 8 and 16 output channels, respectively, and a $5\times5$ max-pooling layer. Model updates are performed with SGD with momentum for 10 iterations only, using a learning rate of 0.01 and a high momentum of 0.9 to not deviate too much from the previous model parameters. The classifier with parameters $\theta$ is trained using a binary cross entropy loss.

\paragraph{Datasets.}
We performed experiments on four long-term datasets, namely \emph{UAV20L}~\cite{mueller2016benchmark}, \emph{YouTubeLong}, \emph{OxUvA}~\cite{valmadre2018oxuva} and VOT2018 long-term \cite{Kristan2018a}. Here, \emph{YouTubeLong} is a set of very long videos (around 25-30 minutes length each) that we gathered from YouTube, sparesely annotated every 25 frames. Note that the videos are annotated sparsely every 100 frames, as this suffices in practice~\cite{valmadre2018oxuva}. It also contains ``absent'' annotations. In addition OTB50 dataset has been used for a shot-term tracking study. Note that building a tracker for short-term scenarios is beyond the scope of this paper, however, it is important that any tracker that works well for long-term should as well be acceptable for short-term scenarios.

\paragraph{Evaluation metrics.}
We use the AUC metric as in OTB~\cite{wu2015object} and UAV20L~\cite{mueller2016benchmark} adapted to also incorporate  absent frames: any predicted box on a frame with an ``absent'' annotation is penalized aggressively with the \emph{worst possible IoU}, namely 0.
A detection is successful if the IoU is larger than a threshold, and the percentage of successfully tracked frames is calculated.
We plot the curve for the AUC by varying the IoU threshold. 
As recommended in~\cite{valmadre2018oxuva} for OxUva we use the true positive ration (TPR). For VOT2018 long-term dataset, Precision, Recall and F-score, as defined in \cite{Kristan2018a}, are used.

% We denote the modified AUC metric still as \textit{AUC} for convenience.

\begin{table}
    \renewcommand{\arraystretch}{1.0}
    \centering
    \scalebox{0.85} {
    \setlength{\tabcolsep}{6pt}
    \begin{tabular}{lcc}
        \toprule
        & \textbf{UAV20L} & \textbf{YouTubeLong}  \\
        \midrule
          Local Search & 39.8 & 23.3 \\
          Global Search & 49.4 & 37.6 \\
        \bottomrule
    \end{tabular}
    }
    \caption{Comparison between global search and local search on UAV20L and YouTubeLong, measured in AUC (\%). Global search is advantageous when tracking for long duration where the target might disappear and reappear.}
    \label{tab:global_vs_local}
    \vspace{-4mm}
\end{table} 

\subsection{Evaluating Full Image Search}

We first evaluate what is the effect of global and hybrid search in long-term tracking, while not performing any model updates. 
We show the results in Table~\ref{tab:global_vs_local}. As expected, global search returns considerably better results, as it eliminates model decay caused by sampling drift. In fact, the difference is even more noticeable on the YoutubeLong, where videos are wilder and longer. Also, the performance of the method was studied for several different intervals $T$ for the hybrid search, and it was found that the hybrid search with $T = 15$ permitted real time tracking at 25 fps without any negative impact on tracking accuracy.

\subsection{Evaluating the Decay Recognition Network}

Next, we evaluate the decay recognition network and its effect on model decay in long-term tracking.
We compare four variants: global search tracking with LT-SINT with \emph{(i)} no model updates at all, \emph{(ii)} with model updates in  every frame, \emph{(iii)} using a siamese $0.5$ similarity threshold as a decay recognition network and \emph{(iv)} using the LSTM decay recognition network.
We show results in Table~\ref{tab:sa_upd}.

We observe that the LSTM decay recognition network is able to minimize model decay in both UAV20L and YoutubeLong.
An interesting observation is that on the hard and long YouTubeLong videos updating at every frame causes a large drop in accuracy because of the large model decay.
We conclude that the LSTM decay recognition network can predict well when an update is likely to cause contribute bias to the model, while allowing for adapting to the changing appearance of the target objects.
% \footnote{More evaluation of the self-aware update and the self-evaluation module is presented in the supplementary material.}

\begin{table}
    \renewcommand{\arraystretch}{1.0}
    \centering
    \scalebox{0.85} {
    \setlength{\tabcolsep}{6pt}
    \begin{tabular}{lcc}
        \toprule
        & \textbf{UAV20L} & \textbf{YouTubeLong}  \\
        \midrule
          \textbf{LT-SINT with} \\
          no model updates & 49.4 & 37.6 \\
          updates at every frame & 50.4 & 25.3 \\
          siamese similarity updates & 51.4 & 40.4 \\
          LSTM decay recognition network & 53.9 & 41.9 \\
        \bottomrule
    \end{tabular}
    }
    \vspace{-2mm}
    \caption{Comparison between the proposed self-aware model update and three baselines, measured in AUC (\%). ``no-upd'' does not update the model. ``blind-upd'' updates every time. ``sim-upd'' updates the model when the similarity score of the predicted box is above the threshold (0.5). The proposed ``selfaware-upd'' is effective in handling model drift, achieving the best performance.}
    \label{tab:sa_upd}
    \vspace{-3mm}
\end{table} 

\subsection{State-of-the-Art Benchmarking}

\paragraph{Long videos.} We compare the proposed Long-term Siamese Tracker with other state-of-the-art methods on the UAV20L, on YouTubeLong, OxUva and VOT2018 long-term datasets.

\emph{UAV20L, YouTubeLong, OxUva.} Results related to these datasets are presented in Tables~\ref{tab:state_of_the_art}. It is observed that ECO-DEEP, although being a short-term tracker, works well on UAV20L even for videos up to 100 seconds long. The reason is that UAV20L videos undergo mild appearance changes of the targets. On the half-an-hour hard videos in YouTubeLong, however, ECO-DEEP is less successful, as several risky updates lead to model decay.
And, on the diverse and very long OxUva videos LT-SINT has $10\%$ margin from the second best tracker.
To evaluate if the various trackers can locate the target till the end, we examine the last 20 seconds of the YoutubeLong videos, where LT-SINT maintains an excellent accuracy. We conclude that LT-SINT achieves top performance in long-term tracking for these datasets. 

\emph{VOT2018 long-term.} On this dataset, LT-SINT provides an F-score of 0.536, which is a bit behind some of the state-of-art trackers. However, this should not be counted as a limitation of LT-SINT. Among the state-of art trackers, SiamRPN++ uses a deep network architecture, namely ResNet50, DaSiamLT uses a region proposal framework and MBMD employs a SSD-MobileNet architecture. Clearly, the use of these frameworks and larger training datasets on top of a plain Siamese network improves the predictions and leads to higher values of F-score. Since the goal of this paper is to present a theoretical analysis of model bias and pave way for research to reduce it, we limited ourselves to trackers based on basic Siamese matching. In this context, a fair comparison would be with SiamFC, and we see that adding the decay recognition network already helps in improving the performance of our tracker.

\textbf{Short videos.} While model decay is visible in long videos, for completeness we evaluate LT-SINT also on the short videos of OTB~\cite{wu2015object}, using the standard AUC metric.
Without exhaustive hyperparameter optimizations, LT-SINT scores $59.8$ AUC, which is in the same ballpark as other popular short-term trackers that rely on aggressive model updating, like ECO~\cite{danelljan2017eco} ($69.1$), CFNet~\cite{valmadre2017end} ($58.6$) or SiamFC~\cite{bertinetto2016fully} ($58.2$). The slightly lower performance on short term videos can be expected by long-term trackers as in effect, long-term suited trackers have an additional constraint.

\begin{table}
    \renewcommand{\arraystretch}{1.0}
    \centering
    \scalebox{0.8} {
    \setlength{\tabcolsep}{6pt}
    \begin{tabular}{lcccc}
        \toprule
        & \textbf{UAV20L} & \textbf{YouTube} & \textbf{YouTube}  & \textbf{OxUvA}  \\
        &  & \textbf{Long} & \textbf{Long} \emph{(20 sec)} &  \\
        \midrule
          ECO-DEEP~\cite{danelljan2017eco} & 42.7 & 7.1 & 1.4 & 39.5 \\
          TLD~\cite{kalal2012tracking} & 22.8 & 22.4 & 20.2 & 20.8\\
          LTCT~\cite{MaYZY15} & 25.5 & 2.2 & 0.2 & 29.2 \\
          SPL~\cite{supancic2013self} & 35.6 &  \textemdash & \textemdash & \textemdash \\
          SRDCF~\cite{Danelljan2015} & 34.3 & \textemdash & \textemdash & \textemdash \\
          MUSTer~\cite{Hong2015} & 32.9 & \textemdash & \textemdash & \textemdash \\
          
          SiamFC+R~\cite{valmadre2018oxuva} & \textemdash & \textemdash & \textemdash & 42.7 \\
          SiamFC~\cite{valmadre2017end} & \textemdash & \textemdash & \textemdash & 39.1 \\
          MDNet~\cite{nam2016learning} & \textemdash & \textemdash & \textemdash & 47.2 \\
          SINT~\cite{tao2016sint} & 49.4 & 37.6 & \textemdash & 42.6 \\
          EBT~\cite{Zhu_2016_CVPR} & \textemdash & \textemdash & \textemdash & 32.1 \\
          BACF~\cite{Galoogahi2017LearningBC} & \textemdash & \textemdash & \textemdash & 31.6 \\
          Staple~\cite{Bertinetto_2016_CVPR} & \textemdash & \textemdash & \textemdash & 27.3 \\
          \hline
          LT-SINT & \textbf{52.4} & \textbf{42.1} & \textbf{39.5} & \textbf{57.9} \\
        \bottomrule
    \end{tabular}
    }
    % \vspace{-2mm}
    \caption{State-of-the-art comparison, measured in AUC (\%). LT-SINT outperforms other trackers on all four datasets.}
    \label{tab:state_of_the_art}
     \vspace{-3mm}
\end{table}

\begin{comment}
\begin{table}
    \renewcommand{\arraystretch}{1.0}
    \centering
    \scalebox{0.8} {
    \setlength{\tabcolsep}{6pt}
    \begin{tabular}{lrrr}
        \toprule
        & \textbf{F-Score} & \textbf{Precision} & \textbf{Recall} \\
        \midrule
        SiamRPN++ \cite{Li2019siamrpnpp} & 0.629 & - & - \\
        MBMD \cite{Kristan2018a} & 0.610 & 0.634 & 0.588 \\
        DaSiamLT \cite{Kristan2018a} & 0.607 & 0.627 & 0.588 \\
         LT-SINT & \textbf{0.536}  & \textbf{0.566} & \textbf{0.510} \\
         SYT & 0.509 & 0.520 & 0.499 \\
         SLT~\cite{Kristan2018a} & 0.456 & 0.502 & 0.417 \\
         SiamFC~\cite{valmadre2017end} & 0.433 & 0.636 & 0.328 \\
         
        \bottomrule
    \end{tabular}
    }
    % \vspace{-2mm}
    \caption{List of trackers used on VOT2018 long-term dataset along with their performance scores.}
    \label{tab:votlt2018}
     \vspace{-3mm}
\end{table}
\end{comment}

\section{Conclusion}
This paper highlights the negative impact that adverse bounding box predictions have to the tracker models, introducing an unvoidable bias term to the learning dynamics of the tracker and leading to model decay.
We examine empirically the effect the various tracking challenges have on model decay.
Based on the insights we propose two simple modifications to siamese trackers, such that model updates are possible without leading to significant model decay.
Experiments on four long-term and one short-term tracking benchmarks show that the proposed tracker is accurate and does not suffers from model decay even in wild, 30-minute long videos.

{\small
\bibliographystyle{ieee}
\bibliography{main_ref}
}

\end{document}